\documentclass[runningheads,a4paper]{llncs}

\usepackage{graphicx}
\usepackage{amsmath,amssymb}
\usepackage{bm}
\usepackage{booktabs}
\usepackage{multirow}
\usepackage{subcaption}
\usepackage{hyperref}
\usepackage{cite}
\usepackage[T1]{fontenc}
\usepackage{xcolor}
\usepackage[normalem]{ulem}   

\AtBeginDocument{\def\doi#1{}}

\AtBeginDocument{%
  \setlength{\abovedisplayskip}{6pt plus 2pt minus 2pt}%
  \setlength{\belowdisplayskip}{6pt plus 2pt minus 2pt}%
  \setlength{\abovedisplayshortskip}{3pt plus 2pt}%
  \setlength{\belowdisplayshortskip}{4pt plus 2pt}}

\colorlet{delcolor}{gray}

\begin{document}

\title{Multi-Modal Non-Prehensile Estimation of Physical Parameters via Press-and-Pull Tipping}
\titlerunning{Multi-Modal Non-Prehensile Estimation of Physical Parameters}

\author{Steven M. Hyland, Jing Xiao, and Cagdas D. Onal}
\authorrunning{Hyland, Xiao, and Onal}
\institute{Worcester Polytechnic Institute \\ \email{smhyland@wpi.edu, jxiao2@wpi.edu, cdonal@wpi.edu}}

\maketitle

\begin{abstract}
Recovering physical properties of unknown objects through non-prehensile interaction is challenging because no single manipulation primitive reveals all relevant parameters. Planar pushing couples mass and friction, while conventional tipping cannot recover friction and may fail entirely when low-friction or curved-base objects slide or rotate instead of tipping. We introduce a multi-modal estimation framework that combines a sliding interaction with a press-and-pull tipping primitive to recover object mass, center-of-mass height, and surface friction. The press-and-pull interaction increases the object–table sliding threshold and stabilizes the pivot, enabling controlled tipping without any prior geometric object model. Wrist force/torque sensing, RGB-D perception, and robot proprioception are fused to estimate the physical parameters from the two complementary interaction modes. Experiments on an ABB IRB120 across four objects with varied geometry, mass, center of mass, and friction achieve low relative error, while successfully operating on curved-base objects that fail under conventional forward tipping. The results demonstrate that complementary non-prehensile interactions can recover a compact set of physical parameters without grasping, a prior object model, or learned interaction dynamics.
\end{abstract}

\section{Introduction}
\label{sec:intro}

Robots increasingly operate in environments where objects are too large, fragile, or unwieldy to be grasped: lifting a monitor by an edge, nudging a toolbox into position, or rebalancing a teetering package on a shelf. In all of these tasks, safe manipulation hinges on the robot knowing a small set of physical parameters: mass, center of mass (CoM), and the object-surface friction coefficient $\mu_t$. Together, these parameters determine when an object will tip, slide, or remain stable. None of these quantities can be reliably read from vision alone.

A natural alternative is to estimate these parameters \emph{through interaction}. Non-prehensile primitives such as pushing and tipping require no grasp planning, no mechanical fixturing, and no prior geometric model, and they apply to objects that are not easily graspable. No single primitive, however, recovers all three parameters: pushing and tipping are each informative about one part of the parameter set and uninformative about the other, and tipping in particular breaks down for low-friction objects that slide before they tip. We review these limitations in detail in Section \ref{sec:related}.

We exploit the fact that these limitations are complementary: the parameter combination one mode confounds, the other resolves. We also implement a means of precise tipping through the \emph{press-and-pull} interaction.

\subsection{Related Work}
\label{sec:related}

\textbf{Inertial Parameter Estimation Through Interaction}
Inertial parameter identification has a long history in robotics, from rigid-body dynamic identification of manipulators to in-hand estimation of grasped payloads \cite{kubus_-line_2007, sundaralingam_-hand_2021}. Mavrakis and Stolkin \cite{mavrakis_estimation_2020} survey the field along three axes: the assumed contact model, the sensor modality, and whether estimation is passive or active. Most existing techniques assume a stable grasp \cite{sundaralingam_-hand_2021, kubus_-line_2007}, which limits their applicability to objects that cannot be securely held. Our framework operates in the non-prehensile regime, treating the object--table contact as a passive contributor to the wrench balance rather than something to be eliminated.

\smallskip
\noindent
\textbf{Pushing-Based Estimation}
Quasistatic planar pushing has been studied extensively as an active-perception primitive \cite{mason_mechanics_1986, lynch_dynamic_1996, yu_more_2016, bauza_probabilistic_2017}. Some approaches learn pushing dynamics for control but treat the underlying parameters as latent \cite{yu_more_2016}, while others explicitly characterize the identifiability problem: in the quasistatic regime, a small number of pushes provides mass--friction pairs, retaining a probability distribution over both \cite{bauza_probabilistic_2017}. The Force Push controller sidesteps the identifiability issue by closing the loop on force feedback alone, never explicitly recovering parameters \cite{heins_force_2024}. Others fuse vision and tactile sensing within a differentiable filter to actively select informative pushes, achieving joint inference of friction, mass, CoM, and inertia, but require learned object--robot interaction models \cite{dutta_push_2023}. Recent work extends pushing-based estimation to mobile robots \cite{hyland_predicting_2023} and to onboard sensing without motion capture \cite{hyland_onboard_2025}. Our approach is complementary: rather than trying to disentangle $m$ and $\mu_t$ within a single quasistatic interaction, we activate modes sequentially to reveal the parameters each mode is suited for.

\smallskip
\noindent
\textbf{Friction Estimation}
Friction estimation has historically been pursued through sliding interactions \cite{lynch_manipulation_1992, bauza_probabilistic_2017}, lateral pressing with tactile sensing \cite{le_probabilistic_2021}, or slip-detection control loops that command motion until tactile slip is detected and then identify $\mu$ at slip onset \cite{sundaralingam_-hand_2021}. Slippery-terrain methods for legged systems use proprioceptive ground-reaction estimates to infer terrain friction \cite{kim_online_2025}. Our approach borrows the slip-onset principle but transposes it to a non-prehensile setting.

\smallskip
\noindent
\textbf{Tipping and Toppling}
Tipping has been studied as both a manipulation goal in its own right, e.g., for object reorientation \cite{lynch_dynamic_1996}, and as an estimation tool. Early tipping-based estimators rely on explicit shape models or vision-based pose tracking through the toppling threshold \cite{maeda_planning_2005}, both of which fail on transparent or texture-less objects. One similar line of work is estimating CoM using Gravity Equi-Effect Planes \cite{yong_yu_estimation_1999} for basic shape primitives through multiple tip interactions across unique axes. This work was later extended for round-base objects \cite{yong_yu_estimation_2001}, though for both, object shape is known a priori, which our approach is not limited by.

\smallskip
\noindent
\textbf{Multi-Modal and Multi-Contact Estimation}
A growing body of work fuses multiple modalities or interaction modes for estimation. Visuo-tactile pipelines \cite{dutta_push_2023, le_probabilistic_2021} combine RGB-D shape priors with tactile force readings. Robust transport methods \cite{heins_robust_2025} explicitly account for inertial uncertainty rather than estimating it. Differentiable physics-informed world models \cite{li_pin-wm_2025} jointly identify multiple physical parameters through few-shot pushing trajectories. In contrast, our approach is deliberately minimalist: a coordinated multi-modal framework, an analytical wrench-balance model, and decoupled sub-problems with closed-form structure, requiring no learned model and no shape prior.

\subsection{Contributions}
This paper makes the following contributions about estimating physical parameter values of rigid objects:
\begin{itemize}
  \item \textbf{A complementary multi-modal estimation framework.} Sliding observes $\mu_t m$, while pivot tipping recovers $(m,z_c)$ independently of friction. Transforming the measured F/T wrench to the vision-estimated pivot and combining both modes recovers $(m,z_c,\mu_t)$ without explicit fingertip localization, a second sliding interaction, a learned model, or a shape prior.

  \item \textbf{The press-and-pull primitive.} A downward press raises the object--table sliding threshold and pins the pivot while a horizontal pull induces tipping. Adaptive press selection enables low-friction and curved-base objects to tip rather than slide or swirl, without prior friction knowledge.

  \item \textbf{Demonstration across a challenging object range.} Physical validation without per-object tuning across $0.24$--$5.0$\,kg objects with varied geometry and friction, including a curved-base object that defeats standard tipping.
\end{itemize}

\section{Methodology}
\label{sec:methodology}
This section provides background on standard tipping, sliding, and the multi-modal framework proposed by this research.

\subsection{Background: Forward Tipping}
Tipping-based estimation recovers $(m, z_c)$ from the torque 
balance at a fixed pivot edge $\{O\}$ \cite{hyland_before_2026, yong_yu_estimation_1999}:
\begin{equation}
  \left( ^O\mathbf{p}_{push} \times \! ^O\mathbf{F}_{push} \right) + \left( ^O\mathbf{p}_c \times \!^O\mathbf{F}_{grav} \right) = \mathbf{0},
  \label{eq:torque-balance}
\end{equation}
from which $(m, z_c)$ follow while $\mu_t$ drops out entirely: tipping is friction-agnostic by construction. Two limitations remain unresolved. First, $\mu_t$ is unrecoverable from any pure tipping interaction. Second, for smooth or low-friction objects, sliding precedes tipping for all admissible push geometries, making $(m, z_c)$ unrecoverable as well. This paper resolves both.

\smallskip
\noindent
\textbf{Why Pure Tipping Fails for Smooth Objects}
A limitation not addressed by \cite{hyland_before_2026} and \cite{yong_yu_estimation_1999} is \emph{tipping feasibility}: whether the object is guaranteed to tip without slipping, often maximized by pushing at the top of the object.

Recall the sliding and tipping constraints for an object with arbitrary applied force $\mathbf{F}_{app}$ acting at point $\mathbf{p}_{app}$, \cite{lynch_modern_2017}:
\begin{equation} \label{eq:tip-slide}
    F_{slide} = \mu_t N 
    \qquad \qquad 
    F_{tip} = \frac{\|\mathbf{p}_c \times \mathbf{F}_{grav}\|} {(\|\mathbf{p}_{app}\| \sin\alpha)} \, ,\quad \alpha : \angle(\mathbf{p}_{app}, \mathbf{F}_{app})
\end{equation}

The object begins at rest and force is applied gradually, and the relative ordering $F_{tip} \gtrless F_{slide}$ of these two thresholds determines the resulting motion. 

For smooth-bottomed objects, $F_{slide} < F_{tip}$ may hold for \emph{all} admissible push heights, so sliding always precedes tipping. In this regime no choice of push geometry can recover $(m, z_c)$. 

\begin{figure}[tb]
  \centering
  \includegraphics[width=0.85\linewidth]{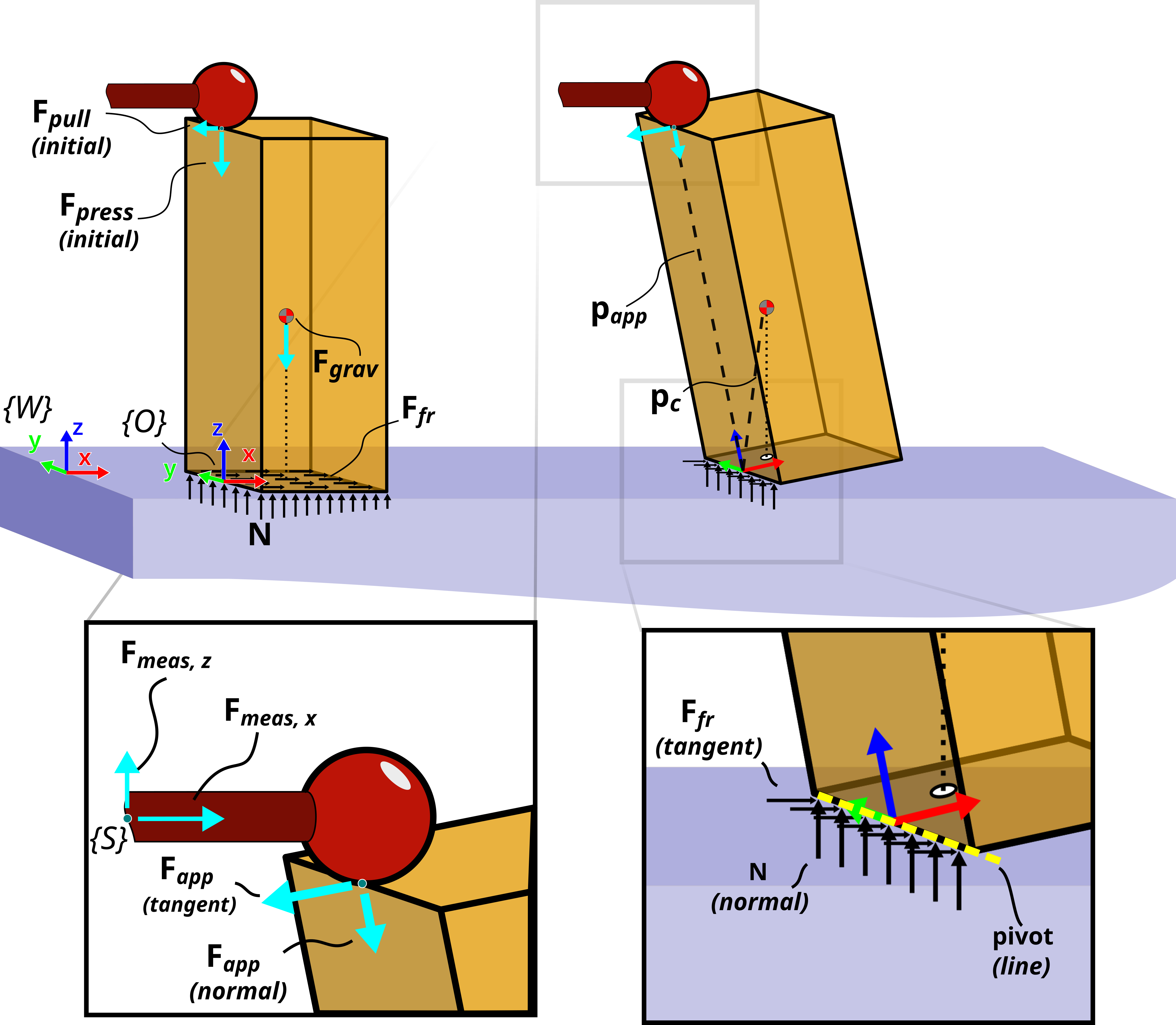}
  \caption{The press-and-pull primitive. A downward press raises the sliding threshold at the object--table interface, while a horizontal pull induces tipping about the near-robot pivot edge.}
  \label{fig:presspull}
\end{figure}

\subsection{Mode 1: Sliding for Friction Estimation} \label{sec:sliding}

To identify the object surface friction coefficient $\mu_t$, the sliding mode probes the force balance at slip onset, where it observes the coupled product $\mu_t \cdot m$.
 
The fingertip approaches the object at a low height, and
moves at low constant speed while F/T measurements are taken. Guaranteed sliding is non-trivial, as well: some objects have high enough friction that sliding without tipping is infeasible. In such cases, Mode 1 cannot recover a unique friction coefficient; it can at most establish a lower bound corresponding to the largest tangential force experienced without slip. This basic manipulation primitive is well-documented in the literature \cite{mason_mechanics_1986, lynch_manipulation_1992, yu_more_2016, zhong_activepusher_2025}.

\subsection{Mode 2: Press-and-Pull for CoM and Mass Estimation} \label{sec:presspull}
To enforce tipping over sliding, we propose a contact motion primitive called \emph{press-and-pull}, illustrated in Figure \ref{fig:presspull}. Satisfying the tipping constraint $F_{tip} < F_{slide}$ admits two complementary strategies: \textit{decreasing} $F_{tip}$ or \textit{increasing} $F_{slide}$ as highlighted by Eq. \eqref{eq:tip-slide}. The robot has affordance over the finger contact point $\mathbf{p}_{app}$ and, to some extent, the applied force $\mathbf{F}_{app}$. In order to minimize $F_{tip}$, the tipping moment arm must be maximized by raising $\mathbf{p}_{app}$, which is bounded by the object height. Conversely, to maximize $F_{slide}$ demands raising friction $\mu_t$ (such as changing the support surface) or increasing the normal force $N$. We favor the latter, since the joint selection of $\mathbf{p}_{app}$ and applied force modulates the support normal force directly:

\begin{description}
  \item[Press:] a downward force $F_{press}$ on the top face of the object raises normal force at object-table contact $N \text{=} mg + F_{press}$, and consequently elevates the sliding threshold to $F_{slide} = \mu_t(mg + F_{press})$.
  \item[Pull:] a horizontal force $F_{pull}$ is the primary driver of tipping about the near pivot edge of the object-surface contact.
\end{description}

This primitive also biases the pivot toward the near-robot side of the object, where it is typically more observable than the far side. When a feasible $F_{press}$ satisfying $F_{tip} < F_{slide}$ exists, tipping is favored over sliding without requiring prior knowledge of $\mu_t$. The sustained press additionally increases frictional engagement at the object-table interface, stabilizing the pivot throughout the motion. This is particularly useful for curved-base objects, for which the active contact can otherwise migrate during tipping. 

\smallskip
\noindent 
\textbf{Interaction Trajectory Generation}
Assuming no-slip contact at the fingertip and table, the pull phase follows a circular arc trajectory centered at the estimated pivot $\{O\}$. The fingertip moves along this arc while maintaining the commanded downward press force, so that the object rotates quasistatically about the pivot rather than translating across the table. 

The finger orientation remains world-constant, which introduces a traveling contact patch on the spherical fingertip that is difficult to characterize due to fingertip compliance. An alternate strategy is to rotate the finger along with the object angle, which instead reduces the workspace since the wrist joint rapidly approaches the table as tipping progresses. The world-fixed strategy produced sufficiently accurate estimates in preliminary testing and was therefore adopted throughout this study.

\smallskip
\noindent
\textbf{Adaptive Press Selection}
Selection of $F_{press}$ should satisfy three constraints:
\begin{enumerate}
    \item $F_{tip} < F_{slide}$ with a margin sufficient to absorb model uncertainty.
    \item $|\mathbf{F}_{press}|$ should be sufficient to maintain \emph{no-slip} between finger and object.
    \item $|\mathbf{F}_{press}|$ should remain sufficiently small to avoid deformation of the top face and excessive resistive moment about the pivot; the latter depends on the press-contact moment arm  $\mathbf{p}_{app}$.
\end{enumerate}
Since the bounds of these constraints are not identifiable until after the tipping procedure, selecting an appropriate press force is non-trivial.

\begin{figure}[tb]
  \centering
  \includegraphics[width=0.92\linewidth]{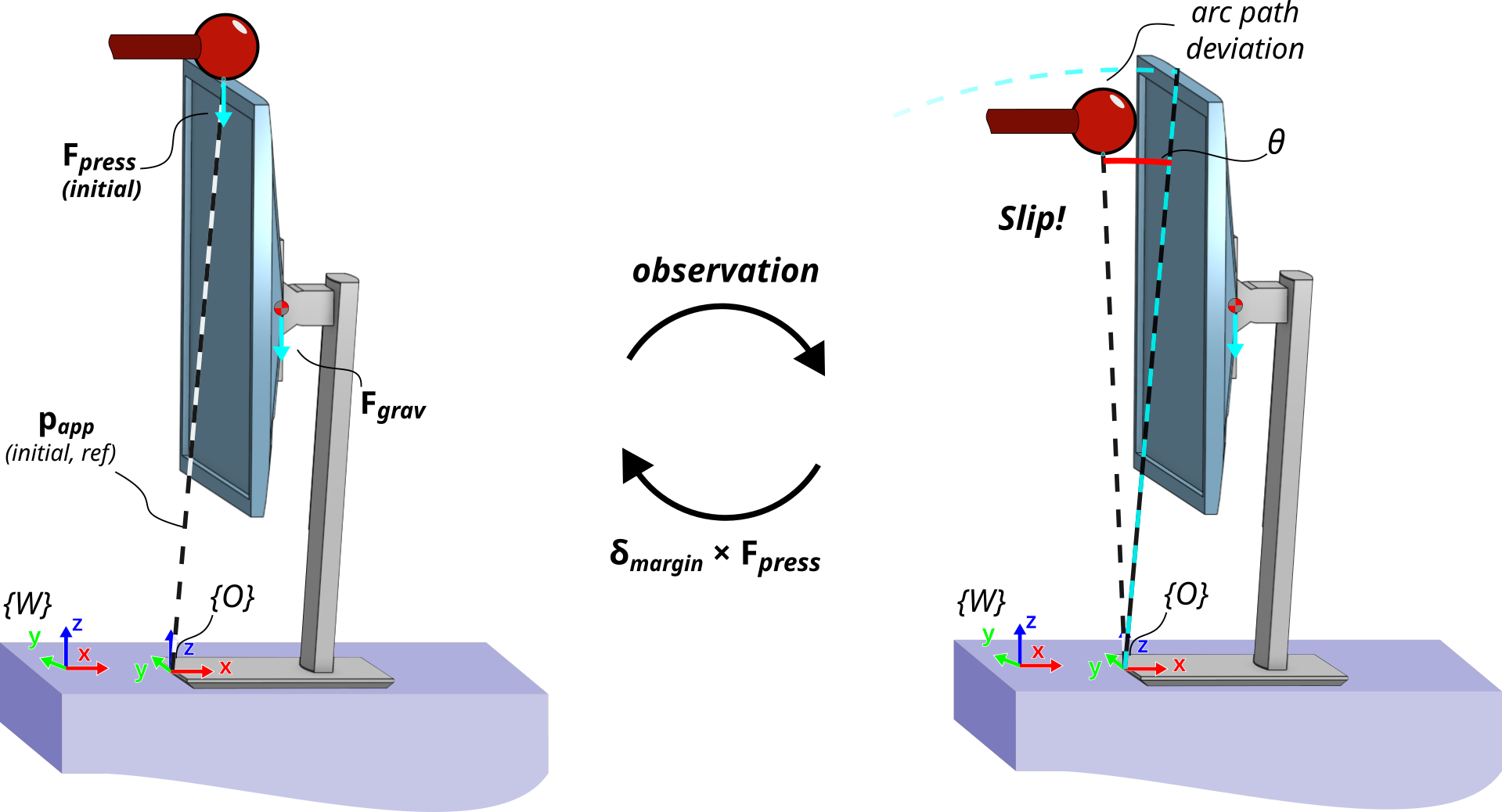}
  \caption{Adaptive press procedure on the monitor object. (right) Slip is detected by deviation from the arc motion and observed vs. expected object pose. Press force that is ultimately found remains constant through the arc for all object angles $\theta$ (from proprioception).}
  \label{fig:adaptive_press}
\end{figure}

Illustrated in Figure \ref{fig:adaptive_press}, we propose an \emph{adaptive press selection} that converges to a press force that satisfies all three constraints. From a small nominal value for $|\mathbf{F}_{press}|$, force/torque measurements and object pose estimate are observed. If the initial press-and-pull fails to produce pure tipping or yields slip at the finger, press force is increased by some margin $\delta _{margin}$, in this case multiplicative. The procedure repeats until clean tipping is observed. The upper bound for $|\mathbf{F}_{press}|$ is set by the shear strength of the rigid finger, though other bounds may be substituted, such as robot joint limits. In practice, we initialize $F_{press} = 5\,\mathrm{N}$ and augment by $\delta_{margin} = 1.25$ per attempt.

\smallskip
\noindent
\textbf{Hysteresis Cancellation}
The push--retract friction cancellation strategy of \cite{hyland_before_2026} carries over: during press-and-pull, the return-to-rest reverses the friction direction at the finger--object contact, producing a hysteresis structure in the measured torque--angle trajectory. Combining both segments in the regression cancels frictional bias in the torque fit.

\begin{figure}[tb]
  \centering
  \includegraphics[width=0.85\linewidth]{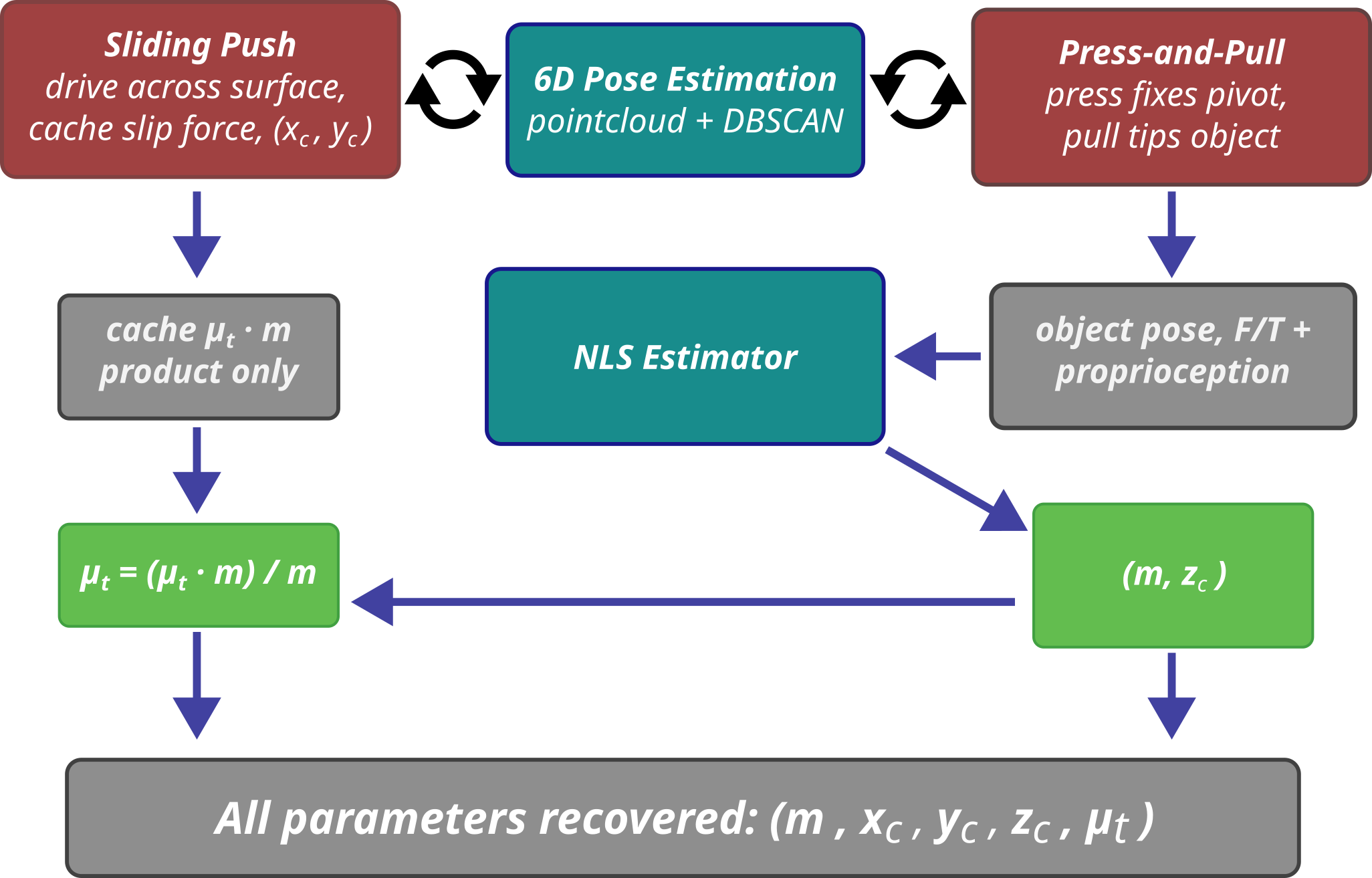}
  \caption{Multi-modal pipeline. Pose estimation is used to detect finger slip and assist with contact-point selection. Friction estimation relies on output from NLS (non-linear least squares) estimation.}
  \label{fig:flowchart}
\end{figure}

\subsection{Inertial Parameter Estimation Model}
Because the pivot is fixed by frictional engagement, a quasistatic wrench-balance analysis is employed. The applied force now carries a large vertical (press) component, so we work directly with the full wrench. Reading the F/T sensor wrench as ${}^{S}\mathbf{w}_{\mathrm{meas}} = [{}^{S}\mathbf{f}_{\mathrm{meas}},\ {}^{S}\boldsymbol{\tau}_{\mathrm{meas}}]^{T}$, we map it to the object frame $\{O\}$ via the adjoint transform to yield the \emph{applied wrench at \{O\}}:
\begin{equation}
  {}^{O}\mathbf{w}_{app} = - \text{Ad}_{T_{SO}}\, {}^{S}\mathbf{w}_{\text{meas}} =
  -\begin{bmatrix} {}^{O}\mathbf{R}_{S} & \mathbf{0} \\ [{}^{O}\mathbf{p}_{S/O}] {}^{O}\mathbf{R}_{S} & \,{}^{O}\mathbf{R}_{S} \end{bmatrix}
  \begin{bmatrix} {}^{S}\mathbf{f}_{\text{meas}} \\ {}^{S}\boldsymbol{\tau}_{\text{meas}} \end{bmatrix}
  \label{eq:adjoint}
\end{equation}
where ${}^{O}\mathbf{R}_{S}$ and ${}^{O}\mathbf{p}_{S/O}$ are the orientation (matrix) and position of the sensor frame in $\{O\}$, and $[\,\cdot\,]$ is the skew-symmetric (cross-product) matrix. The rotation and position of the sensor (encoded in transform matrix $T_{SO}$), follow from robot proprioception together with pivot location, known from the intersection of the object convex hull and table plane.

A subtle but important consequence of Eq. \eqref{eq:adjoint} is that the adjoint transform propagates the full measured wrench, directly to $\{O\}$ using only the sensor pose $T_{SO}$. This bypasses the need for explicit contact point calculation (i.e., $\mathbf{p}_{app}$ in $\{O\}$), which would otherwise require extrapolation from the robot kinematic chain and object pose estimate. Since $T_{SO}$ is derived directly from proprioception, this formulation is more accurate than relying on contact point localization.

The complete quasistatic equilibrium at the pivot gives:

\begin{equation}
  {}^{O}\mathbf{w}_{app} + {}^{O}\mathbf{w}_{grav} + {}^{O}\mathbf{w}_{gnd} = \mathbf{0},
  \label{eq:wrench-eq-tip}
\end{equation}
with
\begin{equation}
  {}^{O}\mathbf{w}_{grav} = \begin{bmatrix} m\, {}^{O}\mathbf{g}(\theta) \\ {}^{O}\mathbf{p}_c \times \left(m\, {}^{O}\mathbf{g}(\theta)\right) \end{bmatrix}_{6\times 1}, \qquad
  {}^{O}\mathbf{w}_{gnd} = \begin{bmatrix} [\mathbf{f}_t]_{2\times 1} \\ N_{t} \\ [0]_{3\times 1} \end{bmatrix}_{6\times 1}
  \label{eq:wrenches}
\end{equation}
where $\theta$ is the object out-of-plane rotation (not about table $\hat{\mathbf z}$), and $[\mathbf{f}_t]_{2\times 1}$ and $N_t$ are the friction and normal force at table contact, respectively. The ground wrench has zero moment since all ground reaction forces act through the pivot edge $\{O\}$. The moment rows of Eqs. \eqref{eq:wrench-eq-tip} and \eqref{eq:wrenches} are therefore independent of $\mu_t$ and contain only the unknowns $m$ and $z_c$, which we recover via the tipping and nonlinear least-squares pipeline.

\smallskip
\noindent
\textbf{Friction Estimation}
The task is to determine the friction coefficient, $\mu_t$. Applying the translational force balance to the Mode 1 sliding interaction, the table reaction balances the applied and gravitational load; resolving it onto the table-tangent and table-normal directions gives the friction and normal contact forces
\begin{equation}
  f_t = - \left| \left( {}^O\mathbf{f}_{app} + m\,{}^O\mathbf{g}(\theta)\right)\cdot \hat{\mathbf{v}}_t \right|,
    \qquad
    N_{t} = -\left| \left({}^{O}\mathbf{f}_{app} + m\,{}^O\mathbf{g}(\theta) \right) \cdot \hat{\mathbf{n}}_t \right|,
  \label{eq:slide-balance}
\end{equation}
where $\hat{\mathbf{n}}_t$ is the table normal and $\hat{\mathbf{v}}_t$ is the in-plane unit vector along the direction of incipient slide. Once mass $m$ has been recovered from the tipping mode, the table normal reaction $N_t$ can be computed from force balance. During Mode 1, the sliding interaction produces a measurable tangential reaction $f_t$ that reaches the Coulomb limit at slip onset \cite{lynch_modern_2017}. Thus, $\mu_t$ is made observable, and follows directly as $\mu_t = |f_t|/N_t$. Note that the slip measurement only observes the product of mass and friction, which cannot separate $\mu_t$ from $m$, one of the limitations resolved our multiple interaction modes.

\subsection{Estimation Process} \label{sec:estimation}
As illustrated in Figure \ref{fig:flowchart}, parameter recovery proceeds through two sequential interactions:
\begin{enumerate}
  \item \textbf{Planar push}: drive the object across the surface at low height. Cache the slip-onset tangential force, which observes the product $\mu_t \cdot m$. The same interaction also yields planar CoM components $(x_c, y_c)$ via planar pushing techniques \cite{hyland_predicting_2023}.
  \item \textbf{Press-and-pull tipping}: press fixes the pivot; a slow pull tips the object sub-critically. Torque balance recovers $(m, z_c)$ and the cached product then yields $\mu_t=f_t \,/\, mg$.
\end{enumerate}

Both interactions share the following execution protocol:
\begin{enumerate}
  \item Segment the pointcloud and select a contact point appropriate for the interaction mode.
  \item Move the robot to a pre-contact pose near the calculated contact point.
  \item Translate the finger at constant velocity; record the full wrench trajectory.
  \item Detect slip or tip onset; verify no finger slip occurred. Return object to rest.
  \item Apply a median plus Savitzky--Golay filter cascade to suppress sensor noise.
  \item Estimate parameters according to interaction mode using filtered data.
\end{enumerate}

\section{Experiments}
\label{sec:experiments}
\begin{figure}[tb]
  \centering
  \includegraphics[width=0.92\linewidth]{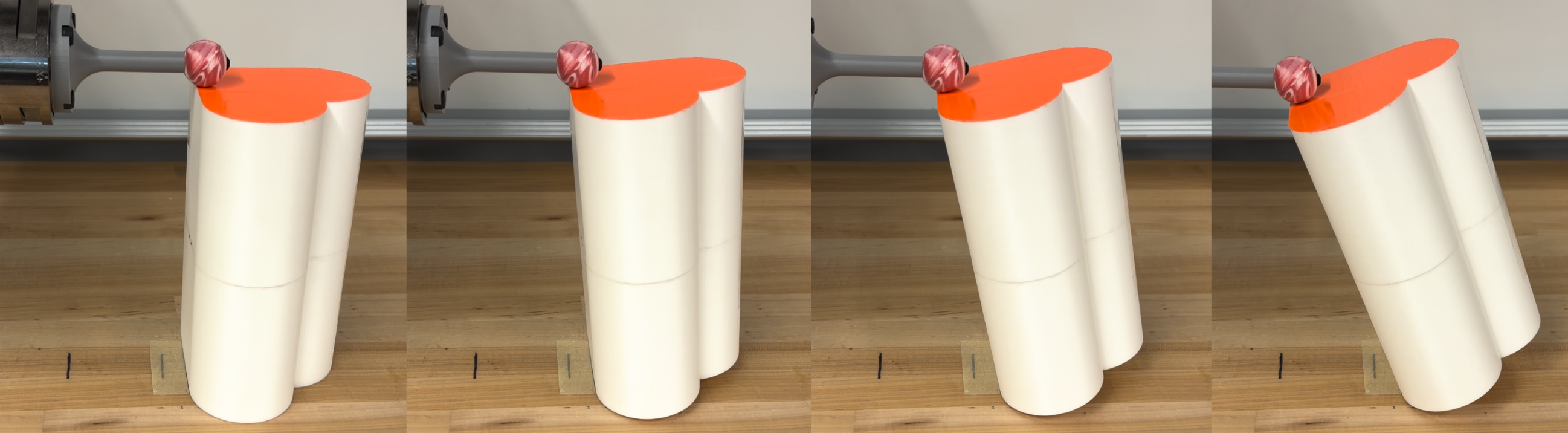}
  \caption{Heart object during the press-and-pull mode. The finger first applies a downward press force. Upon reaching the press force, the finger retracts and applies a pull force, which continues while the finger maintains the press force throughout tipping. This continues until the measured tangential force falls to the defined safety margin $\eta_{safety}$.}
  \label{fig:experiment}
\end{figure}

\begin{figure}[tb]
  \centering
  \includegraphics[width=0.84\linewidth]{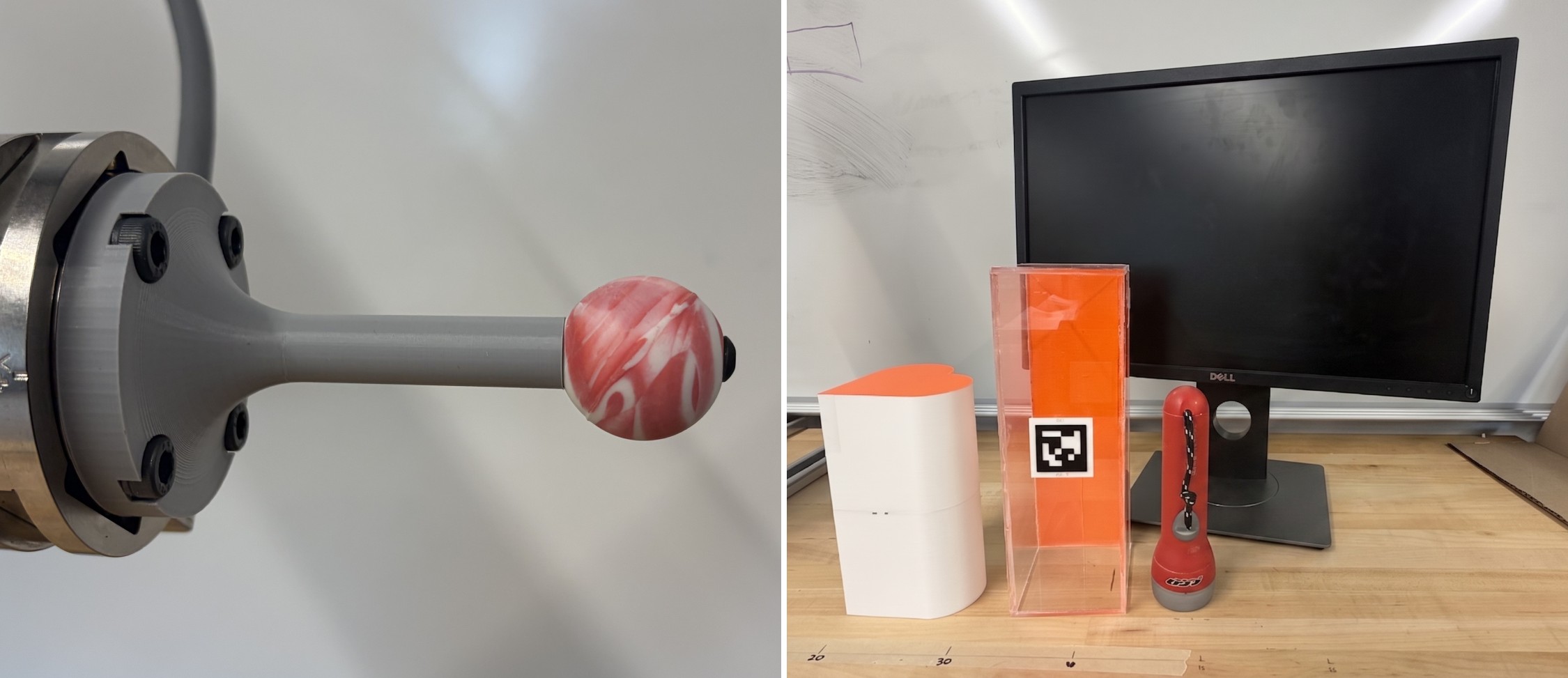}
  \caption{(left) Robot finger used for both interaction modes. High-friction spherical tip provides better traction during press-and-pull and reduces required press force magnitude. (right) four objects tested, varying in size, shape, weight, friction, and CoM.}
  \label{fig:lineup_and_finger}
\end{figure}

Physical experiments use an ABB IRB120 industrial manipulator with a rigid 3D-printed finger with a high-friction, compliant, spherical fingertip which aids the press-and-pull mode (Figure \ref{fig:lineup_and_finger}). Figure \ref{fig:experiment} shows the heart object under the primitive. A wrist-mounted ATI Gamma six-axis Net F/T sensor measures the full interaction wrench at $500\,\mathrm{Hz}$. Object pointclouds are recovered from dual Intel RealSense D435 depth cameras: one set low on the pivot side of the workspace, and one aimed down at the top faces of the objects. All computation is performed on an Intel Core i7-13700H processor.

\subsection{Object Pose Estimation}
\label{sec:pose-estimation}
Two complementary techniques are implemented, with distinct roles. Vision supplies the geometry needed before contact: the contact coordinates and the pivot used to parameterize the press-and-pull arc trajectory. Vision also provides an independent pose observation used to detect finger slip during adaptive press selection. Proprioception then supplies the object angle throughout tipping, where it is smoother and more accurate than the hull.

\paragraph{Pointcloud pose estimation} Two fixed RGB-D cameras stream pointclouds into the world frame $\{W\}$ through a shared extrinsic calibration. The merged cloud is cropped to the workspace and voxel-downsampled for uniform density. Object clusters are extracted with DBSCAN and the corresponding 3D convex hulls are calculated. The hull is the sole source of the interaction geometry: it yields the push and press point candidates of Sec. \ref{sec:push-point-selection} for both modes, and the pivot edge $\{O\}$ about which the press-and-pull arc trajectory is centered. Its pose is additionally compared against the commanded arc trajectory to detect finger slip during adaptive press selection.

The two cameras provide complementary views of the interaction-relevant regions, improving depth accuracy and reducing occlusion, in particular of the pivot/base geometry and the top/press region of the object. The reconstructed convex hull may remain incomplete where surfaces are unobserved. However, pivot localization depends primarily on the locally observed object-table contact geometry. Residual depth noise, segmentation error, and occlusion near this interface can perturb the estimated pivot location.

\paragraph{Proprioceptive pose estimation} The object angle and pivot axis are estimated from robot proprioception during tipping. Provided no finger slip, the rigid object follows the trajectory enforced by the finger, so the object angle $\theta$ and the realized tipping axis location can be derived from the end-effector pose alone. This increases the accuracy in estimating $\theta$: proprioception is free of both depth noise and the residual pivot-localization error above, and yields the smooth, densely sampled $\theta$ trajectory that the torque regression of Sec. \ref{sec:estimation} requires. Its scope is correspondingly narrow - it is valid only while the finger maintains no-slip contact, which is precisely the condition the vision-based check enforces.

\subsection{Push Point and Press Point Selection} \label{sec:push-point-selection}
\begin{figure}[tb]
  \centering
  \includegraphics[width=0.85\linewidth]{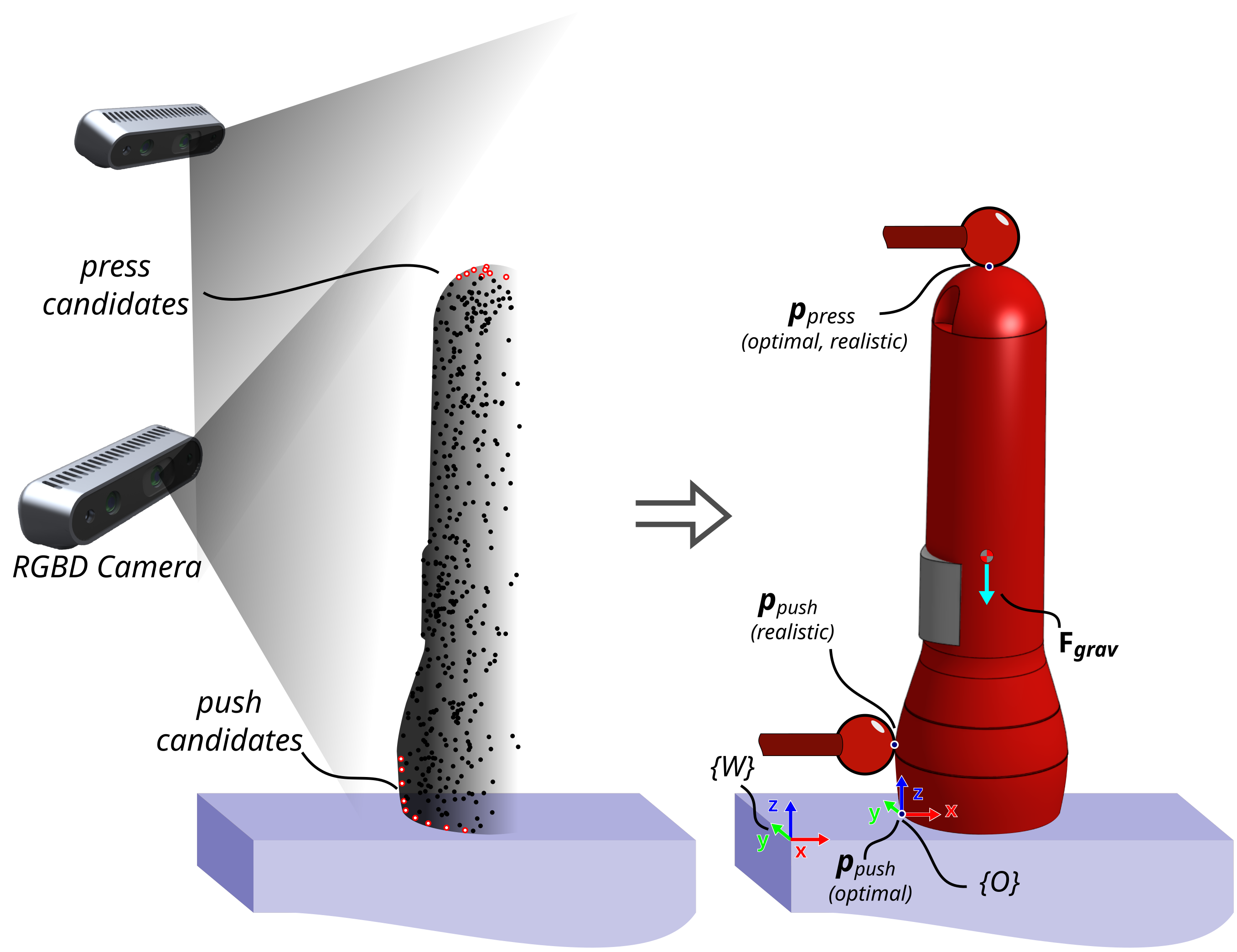}
  \caption{Pointcloud and contact point selection. Candidate points taken from the pointcloud near the extremes of object height, and selected according to feasibility and the heuristics defined in Sec. \ref{sec:push-point-selection}. 
  }
  \label{fig:pointcloud}
\end{figure}

Figure \ref{fig:pointcloud} illustrates the optimal and realistic selection of these contact points and the corresponding pointcloud. 

\paragraph{Push point} The push point selection is relatively simple: choose the lowest-height point of the object on any of its faces that allows free motion of the end effector.

\paragraph{Press point} The press point selection is crucial for stable tipping: select a point on the object top whose face normal approximates world $+\hat z$ (upwards) and minimizes the press force lever-arm ($\min{\mathbf{p}_{app,\, x}}$), often directly above $\{O\}$, so that the press induces minimal torque about $\{O\}$. The pull is then directed towards the pivot edge outward normal.

\subsection{Physical Experiments} \label{sec:physical}
Table \ref{tab:admissible} consolidates the conditions an object must satisfy alongside the component of the pipeline each one serves.

\begin{table}[tb]
  \centering
  \small
  \setlength{\tabcolsep}{3.5pt}
  \caption{Admissible object class: conditions an object must satisfy, and the component of the framework each one serves.}
  \label{tab:admissible}
  \begin{tabular}{l l l}
    \toprule
    Condition & Requirement & Serves \\
    \midrule
    Visibility  & Top face and pivot edge each seen by a camera & Contact selection \\
    Reachability & Top contact and pull arc within the workspace & Press-and-pull \\
    Top friction & Holds finger no-slip under $\mathbf{F}_{press}$ & Adaptive press \\
    Rigidity & Pivot-to-CoM moment arm fixed in $\{O\}$ & Torque balance \\
    Slideability & Slides before tipping under a low push & Mode 1 ($\mu_t$) \\
    \bottomrule
  \end{tabular}
\end{table}

Visibility and reachability are properties of the workcell rather than of the object and are listed only for completeness; any object placed inside the camera baseline and the manipulator's workspace satisfies them. Rigidity is the condition most easily overstated. For example, the monitor's screen visibly flexes relative to its base under load, yet the monitor is among the best performers reported below, because the adaptive press converges to a force at which its deflection stays negligible. We do not quantify how much compliance is tolerable and leave such a study for future work.

Two distinct conditions are top friction and slideability. Top friction gates the press, and is non-trivial precisely because the finger--object friction is unknown prior to contact. This is what the adaptive press loop of Sec. \ref{sec:presspull} resolves. Slideability is required only by Mode 1, and fails only for objects that are extremely easy to tip or are unstable. We observe that this covers the large majority of everyday objects. An object that does violate slideability can still be tipped, so $(m, z_c)$ remain recoverable and only the friction estimate is forfeited. The allowed degree of base roundness is not characterized here and is deferred to future work.

We test our approach on four unique objects shown in Figure \ref{fig:lineup_and_finger}: a hollow acrylic box, a 3D-printed heart-shaped prism, a handheld flashlight, and a computer monitor. Ground-truth $(m, z_c)$ values are taken from independent experiments, while ground-truth $(\mu_t,\theta^*)$ are calculated from mass and CoM values for each object and validated independently.

Finger trajectories follow the protocol of Section \ref{sec:methodology}, with interaction velocities of $5\,\mathrm{mm/s}$ throughout. The arc-motion of press-and-pull continues until the measured pull force reaches the sub-critical safety threshold $\eta_{safety} = 0.1$ of \cite{hyland_before_2026} , i.e., $10\%$ of its peak value at pull onset. The trajectory is then reversed to execute the retract phase symmetrically.

Planar CoM components $(x_c, y_c)$ are an \emph{input} to this framework rather than an output of it: they are supplied by our prior planar-pushing pipeline \cite{hyland_predicting_2023, hyland_onboard_2025}, where they are evaluated separately, and are not re-validated here. They may be recovered from the same Mode 1 sliding push used to cache $\mu_t \cdot m$, so no additional interaction is required.

\begin{table}[tb]
  \centering
  \caption{Estimation results on physical objects. $\hat{m}$ and $\hat{z}_c$ are the mean $\pm$ one SD over $N\text{=}10$ press-and-pull trials; $\hat{\mu}_t$ follows from a single Mode~1 push and the recovered mass.}
  \label{tab:physical-results}
  \small
  \setlength{\tabcolsep}{3pt}
  \begin{tabular}{l | cc | cc | cc}
    \toprule
    & \multicolumn{2}{c|}{$m$ (kg)}
    & \multicolumn{2}{c|}{$z_c$ (cm)}
    & \multicolumn{2}{c}{$\mu_t$} \\
    Object & GT & Estimate & GT & Estimate & GT & Estimate \\
    \midrule
    Box        & 0.676 & $0.707 \pm 0.012$ & 15.00 & $15.10 \pm 0.28$ & 0.156 & $0.149 \pm 0.003$ \\
    Heart      & 0.239 & $0.242 \pm 0.006$ & 10.00 & $11.51 \pm 0.27$ & 0.256 & $0.253 \pm 0.006$ \\
    Flashlight & 0.387 & $0.396 \pm 0.004$ &  9.38 & $9.38 \pm 0.12$ & 0.240 & $0.234 \pm 0.002$ \\
    Monitor    & 5.040 & $5.274 \pm 0.008$ & 23.20 & $24.42 \pm 0.05$ & 0.643 & $0.614 \pm 0.001$ \\
    \bottomrule
  \end{tabular}
\end{table}

\begin{figure}[tb]
  \centering
  \includegraphics[width=0.92\linewidth]{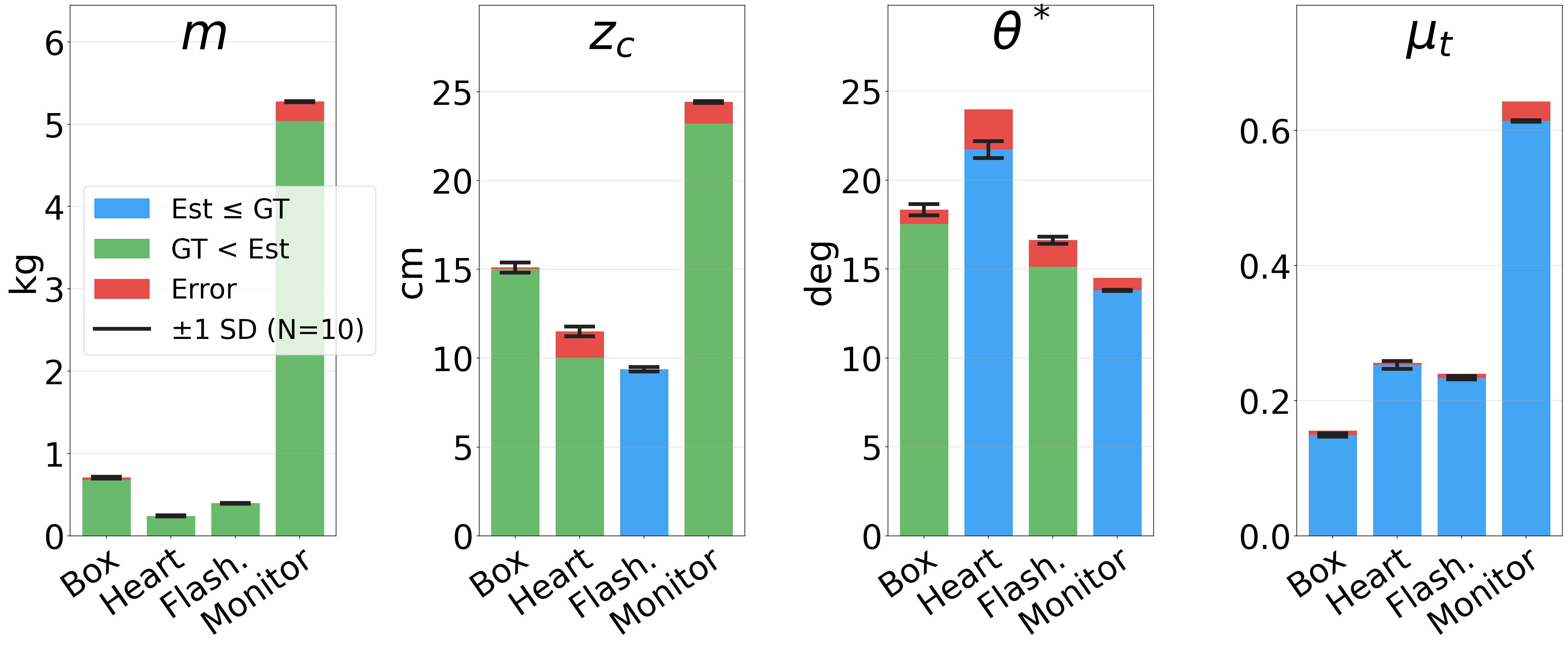}
  \caption{Ground-truth and estimated parameter values across the four objects; error bars show $\pm 1$ SD.}
  \label{fig:abs_error}
\end{figure}

\begin{figure}[tb]
  \centering
  \includegraphics[width=0.92\linewidth]{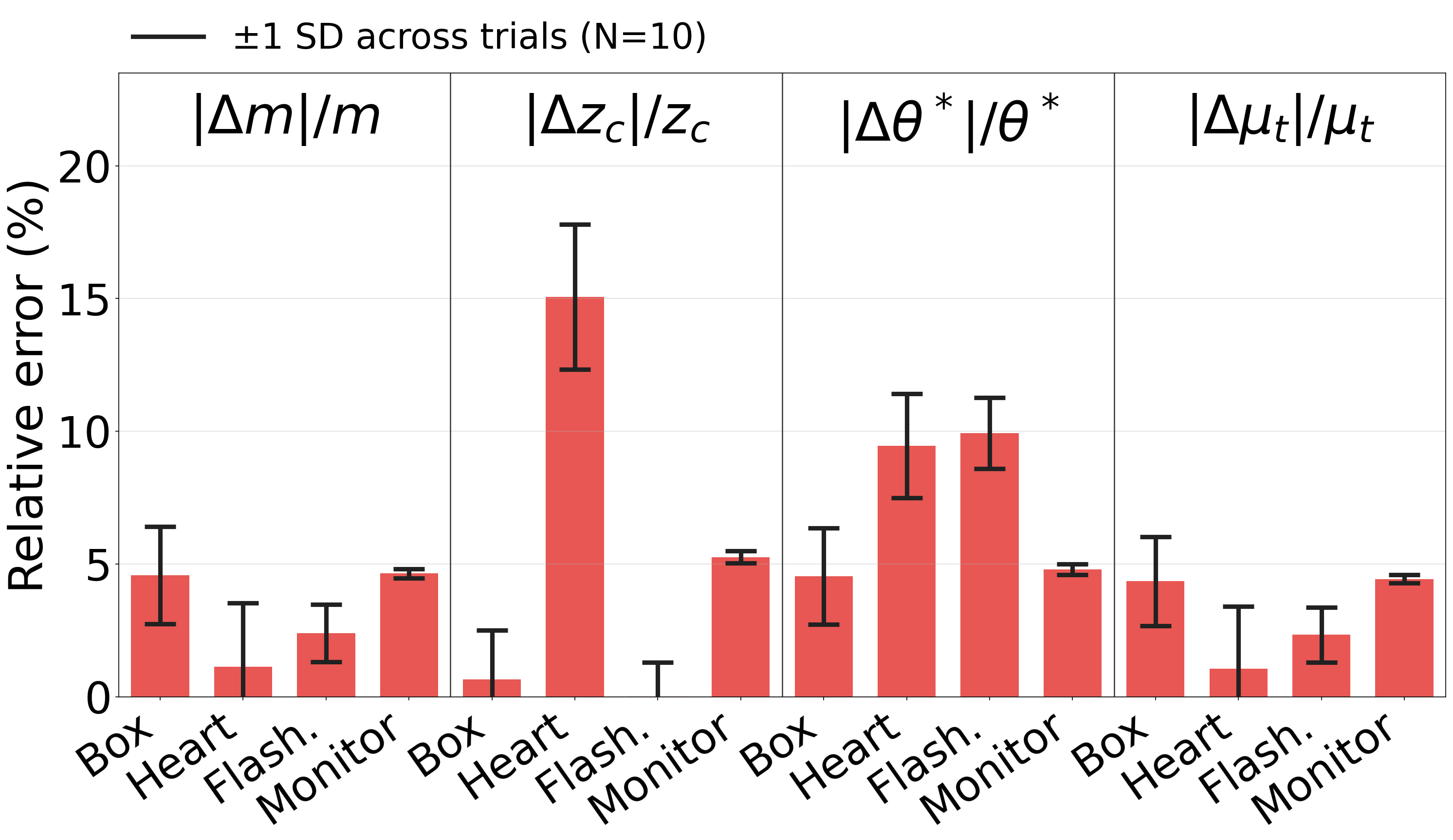}
  \caption{Relative error by object and parameter; error bars show $\pm 1$ SD.}
  \label{fig:rel_error}
\end{figure}



\section{Results and Discussion}

Table \ref{tab:physical-results} summarizes estimation results across the four objects, with absolute errors per parameter in Figure \ref{fig:abs_error} and the corresponding relative errors in Figure \ref{fig:rel_error}. The maximum relative error is $15.1\%$, and three of the four objects remain below $10\%$ across all parameters. We refer the reader to the supplemental video for further information.

\paragraph{Repeatability.}
Each object was tipped $N\text{=}10$ times, and autonomously returned to a nominal rest pose between trials. All 40 press-and-pull trials completed successfully, corresponding to a $100\%$ observed success rate under the study's execution criteria. The estimates show low trial-to-trial dispersion: the coefficient of variation remains below $2.5\%$ for both $m$ and $z_c$ across all objects, and no trial departs from its mean by more than $4.2\%$. Dispersion tracks signal magnitude, the heavier monitor being the tightest (CV $\leq 0.21\%$) and the lightweight heart the widest (CV $\leq 2.48\%$). These statistics characterize empirical consistency across repeated interactions rather than estimator uncertainty; the residual error in Table \ref{tab:physical-results} is correspondingly systematic rather than trial noise.

\paragraph{Estimator consistency.}
Arc and retract phases carry frictional bias with opposite signs, are estimated independently, and are averaged; the two agree to within $\Delta m < 0.26$\,kg, $\Delta z_c < 11$\,mm, and $\Delta\theta^* < 0.63^\circ$ across all rigid objects, confirming that the wrench-balance model is well-conditioned across the tipping range and insensitive to phase direction under the rigid-body assumption. 

\paragraph{Friction estimation.}
Friction errors remain well below $5.0\%$ for every object. The monitor correctly reports the highest $\hat{\mu}_t$, consistent with its rubber feet. Friction error scales with mass error through $\hat{\mu}_t = f_t^{\text{slip}} / (\hat{m}\,g)$, though mass errors are small enough that this amplification remains benign.

\paragraph{Heart object.}
The heart prism produces the single largest error, $15.1\%$ in $z_c$ (Figure \ref{fig:rel_error}).

Two compounding factors degrade estimation quality: its low mass ($239$\,g) produces a small absolute gravitational torque signal, while its geometry requires a comparatively large press force relative to the pull force during tipping. The resulting applied wrench is therefore large relative to the gravitational signal being estimated. The arc/retract estimates agree closely ($\Delta m = 9$\,g, $\Delta z_c = 0$\,mm), and its trial spread, though the widest of the four objects, stays small relative to the estimation error itself, indicating that the observed error is systematic rather than dominated by trial-to-trial noise. Lightweight objects requiring comparatively large press-to-pull force ratios therefore appear to constitute a more challenging regime for the estimator.

\paragraph{Flashlight object.}
The flashlight is a documented failure case for pure forward tipping \cite{hyland_before_2026}: its low-friction curved base causes unwanted swirling about the world $+z$ axis, especially at larger tilt angles near $\theta^*$. The press-and-pull primitive resolves this directly by mechanically pinning the pivot through the press force, and its $m$ and $z_c$ errors are among the lowest of any object (Figure \ref{fig:rel_error}); the CoM height in particular is recovered to within a fraction of a percent.

\paragraph{Monitor object.}
At $5.04$\,kg and a total height of $0.53$ m, the monitor tests estimator robustness at the upper end of the payload range, staying under $6\%$ in every parameter and, by a wide margin, showing the tightest trial-to-trial spread of the four (CV $\leq 0.22\%$). This accuracy is attributed to the heavier monitor weight, yielding a large, clean, and low-noise torque signal during tipping, conditions favored by this estimator.

\section{Conclusions}
\label{sec:conclusion}

We presented a multi-modal non-prehensile framework that recovers an object's mass $m$, center-of-mass height $z_c$, and surface friction $\mu_t$ without a grasp, a shape prior, or a learned interaction model. Sequencing a planar sliding push with a press-and-pull tipping interaction yields all three parameters, and pinning the pivot through a controlled press lets the robot tip smooth- and curved-base objects that standard ``forward'' tipping can only cause to slide or swirl. Across four objects spanning rigid prisms, irregular consumer items, and low-friction contacts, the framework recovered all three parameters with a maximum relative error of $15.1\%$ and resolved the curved-base flashlight, which failed under our prior standard forward-tipping procedure.

The method currently assumes that the top contact is visible and reachable and that the pivot-to-CoM moment arm remains fixed during tipping. Mild compliance was tolerated in the monitor experiments, but articulated or strongly compliant objects may violate this assumption and yield apparent rather than static parameters. In addition, per-trial pivot drift, slip-detection accuracy, and adaptive-press retries were not logged, and the admissible boundary with respect to base roundness has not yet been characterized systematically.

Three natural extensions follow:
\begin{enumerate}
  \item Combining the recovered physical parameters with an object mesh to generate simulation-ready assets for real-to-sim transfer, downstream manipulation planning, and control.
  \item Performing ablations and object sweeps of base roundness and CoM location to characterize the boundary of the admissible object set.
  \item Incorporating an explicit, criticality-aware safety margin into the tipping trajectory, specifically to detect object-table slippage which may occur due to the press force. This would improve safety across all object classes.
\end{enumerate}

\bibliography{references}

\end{document}